\documentclass[11pt,letterpaper]{article}

\usepackage[margin=1in]{geometry}
\usepackage{lmodern}
\usepackage[round,authoryear]{natbib}

\usepackage[utf8]{inputenc}
\usepackage[T1]{fontenc}
\usepackage{hyperref}
\usepackage{url}
\usepackage{booktabs}
\usepackage{array}
\usepackage{graphicx}
\usepackage{amsmath,amssymb}
\usepackage{nicefrac}
\usepackage{microtype}
\usepackage{xcolor}

\title{Patient, Place, Prior (P$^3$): What Counts as Personalization in Medical World Models?}

\author{%
    \normalsize Xingrui Gu\textsuperscript{1}\quad
    Hanxue Gu\textsuperscript{2}\quad
    Yuxiang Zhang\textsuperscript{3}\quad
    Yang Yang\textsuperscript{2}\\[0.5em]
    \small \textsuperscript{1}University of California, Berkeley\\
    \small \textsuperscript{2}University of California, San Francisco\\
    \small \textsuperscript{3}Southern Methodist University
}
\date{}
\hypersetup{
    hidelinks,
    pdftitle={Patient, Place, Prior (P3): What Counts as Personalization in Medical World Models?},
    pdfauthor={Xingrui Gu, Hanxue Gu, Yuxiang Zhang, Yang Yang},
    pdfsubject={Longitudinal medical forecasting and predictive attribution}
}

\begin{document}

\maketitle

\begin{abstract}
Longitudinal models forecast how a patient's imaging state evolves, but accuracy does not show whether the patient's observed trajectory drives the prediction. A population-average forecast may be useful but cannot establish a patient-specific world-model claim. We introduce Patient, Place, Prior (P$^3$), an audit asking whether a forecast benefits from the patient's longitudinal imaging history (Patient), benefits from patient-matched externally supplied spatial support (Place), and gains predictive value beyond a population-average prediction under matched support and context (Prior). We also propose Cancer JEPA, a one-step model that forecasts frozen representations of future breast dynamic contrast-enhanced MRI examinations during neoadjuvant therapy. It adds a lesion-constrained neural correction, trained with an occlusion-based latent objective, to a patient-conditioned low-complexity reduced-rank regression baseline. This factorization permits a post-hoc P$^3$ audit of the frozen model. In a validation cohort previously used in development, forecast error is lower when the neural correction receives the patient's history rather than another patient's and patient-matched lesion occupancy maps rather than substituted maps. However, the descriptive 95\% interval comparing the correction computed from patient history with the population-average neural correction includes zero. P$^3$ thus separates input use from evidence of patient-specific predictive value beyond a population-level pattern.
\end{abstract}

\section{Introduction}

Medical artificial intelligence is expanding from the interpretation of isolated examinations to forecasting changes in patient state. Large-scale self-supervision has produced transferable medical representations \citep{moor2023gmai,zhou2023retfound}, while longitudinal models increasingly address disease progression and individual risk \citep{wang2026retfoundplus}. Serial breast dynamic contrast-enhanced MRI (DCE-MRI) provides an important setting for this transition because it records changes in tumor morphology and enhancement during neoadjuvant therapy \citep{li2020ispy2}. World models offer a framework for representing observations as latent states and learning their transitions \citep{ha2018worldmodels,hafner2019planet,hafner2025dreamerv3}, with emerging medical applications in tumor simulation and clinical digital twins \citep{yang2025mewm,laubenbacher2024digitaltwins}. Applied to medicine, patient-specific forecasting requires predictions that use the patient's observed trajectory in addition to cohort-level dynamics. This motivates our central question: what evidence demonstrates that a medical forecast is patient-specific?

Aggregate predictive accuracy cannot resolve this attribution. Models with comparable held-out performance may rely on different information \citep{damour2022underspecification}, including shortcuts associated with scanner hardware, acquisition protocol, or dataset source \citep{geirhos2020shortcut,degrave2021covid}. Moreover, information encoded about a patient may have little influence on the prediction of interest \citep{brown2023shortcut}. In longitudinal breast MRI, an accurate future-state forecast may derive from the patient's prior scans, an externally supplied lesion mask, or a trajectory shared across the cohort. These sources have different scientific meanings but are collapsed by aggregate error. Saliency maps characterize local sensitivity \citep{arun2021saliency}, while replacing prior scans and lesion masks together confounds their contributions. Evaluating personalization therefore requires matched comparisons that isolate patient history, supplied lesion location, and population-level prediction while preserving the same future target and observed context.

Patient, Place, Prior (P$^3$) provides these comparisons. For patient $i$, \emph{Patient} compares predictions formed from that patient's longitudinal imaging history with predictions formed from another patient's history under the same lesion mask. \emph{Place} compares the patient's lesion mask with another patient's mask while preserving the patient's imaging history. \emph{Prior} compares the prediction formed from the patient's history with a population-average prediction constructed under the same lesion mask and observed context. All comparisons retain patient $i$'s future target and evaluation procedure. Together, they distinguish beneficial use of the patient's imaging history, dependence on supplied spatial support, and predictive value beyond a cohort-level pattern.

We instantiate this framework in Cancer JEPA, a one-step model that forecasts frozen 3DINO representations of future breast DCE-MRI examinations \citep{xu2025threedino}. Following the joint-embedding principle, Cancer JEPA predicts future representations from encoded longitudinal context \citep{lecun2022path,assran2023ijepa,bardes2024vjepa}. It combines a fixed, patient-conditioned reduced-rank regression (RRR) anchor with an occlusion-trained neural correction. Longitudinal imaging history and core--ring lesion occupancy enter the correction separately, and the supplied core occupancy determines where that correction can alter the anchor. Our contributions are: (i) P$^3$, an evidence framework for evaluating patient-specific forecasting; (ii) Cancer JEPA, a spatially constrained latent forecasting model designed for this audit; and (iii) a post-hoc evaluation on longitudinal breast DCE-MRI. Replacing either the patient's imaging history or lesion mask with another patient's increases forecast error, whereas the advantage of patient history over the matched population-average correction remains uncertain. The results therefore distinguish use of patient information from predictive value beyond cohort-level dynamics.

\section{Preliminaries}
\label{sec:preliminaries}

\subsection{Longitudinal latent MRI forecasting}

For patient $i$ imaged at time $t$, let $X_i^t$ denote an acquired MRI volume. An encoder $E$ maps the examination to a spatial representation
\begin{equation}
    Z_i^t = E(X_i^t).
\end{equation}
Self-supervised representations can transfer across medical imaging tasks \citep{litjens2017survey,xu2025threedino}, but $Z_i^t$ may encode scanner hardware and acquisition protocols alongside anatomy and pathology \citep{dinsdale2021unlearning}. We therefore treat it as an encoding of the observed MRI examination, not a direct measurement of biological state.

Given longitudinal imaging history through time $t$, externally supplied spatial support $S_i^t$, the forecast interval $\Delta_i$, and observed acquisition or clinical context $a_i$, a longitudinal model predicts

\begin{equation}
    \widehat{Z}_i^{t+\Delta}
    = F\!\left(C_i^t,S_i^t,U_i\right),
    \label{eq:longitudinal_predictor}
\end{equation}
where $C_i^t$ collects the encoded longitudinal imaging history and $U_i=(\Delta_i,a_i)$ collects the remaining observed context. Longitudinal models use such repeated observations to represent time-varying imaging states \citep{ouyang2022longitudinal}. We study prediction of a future latent MRI state rather than reconstruction of the complete image.

Agreement with a follow-up MRI examination may arise from temporal continuity as well as learned change, for example by copying the current observation or suppressing acquisition noise \citep{gong2024individualised}. We therefore interpret Equation~\ref{eq:longitudinal_predictor} as an observational forecast of an encoded future examination. Without assumptions governing treatment assignment, confounding, and identifiability, it does not identify a causal response to an intervention \citep{prosperi2020causal}.

\subsection{Patient conditioning and predictive personalization}
\label{sec:attribution_problem}

A longitudinal predictor is \emph{patient-conditioned} when it receives the patient's imaging history. Input availability alone does not establish that this information influences the forecast: predictors with similar accuracy can rely on different or unintended signals \citep{damour2022underspecification,geirhos2020shortcut,brown2023shortcut}. We call patient information \emph{beneficially used} when the patient's future is predicted more accurately from that patient's history than from another patient's history under otherwise matched conditions. We reserve \emph{patient-specific predictive value} for the stronger case in which the patient's history also improves on a population-average prediction under the same spatial support and observed context.

Externally supplied spatial support is a separate conditioning source. Improvement with patient-matched support shows that the forecast benefits from the supplied location, whereas localization would require the model to infer that location from the image. Patient conditioning, beneficial information use, patient-specific predictive value, and dependence on supplied spatial support are therefore distinct properties of a longitudinal predictor.

\section{Methodology}
\label{sec:method}

\begin{figure}[t]
    \centering
    \includegraphics[width=0.96\linewidth]{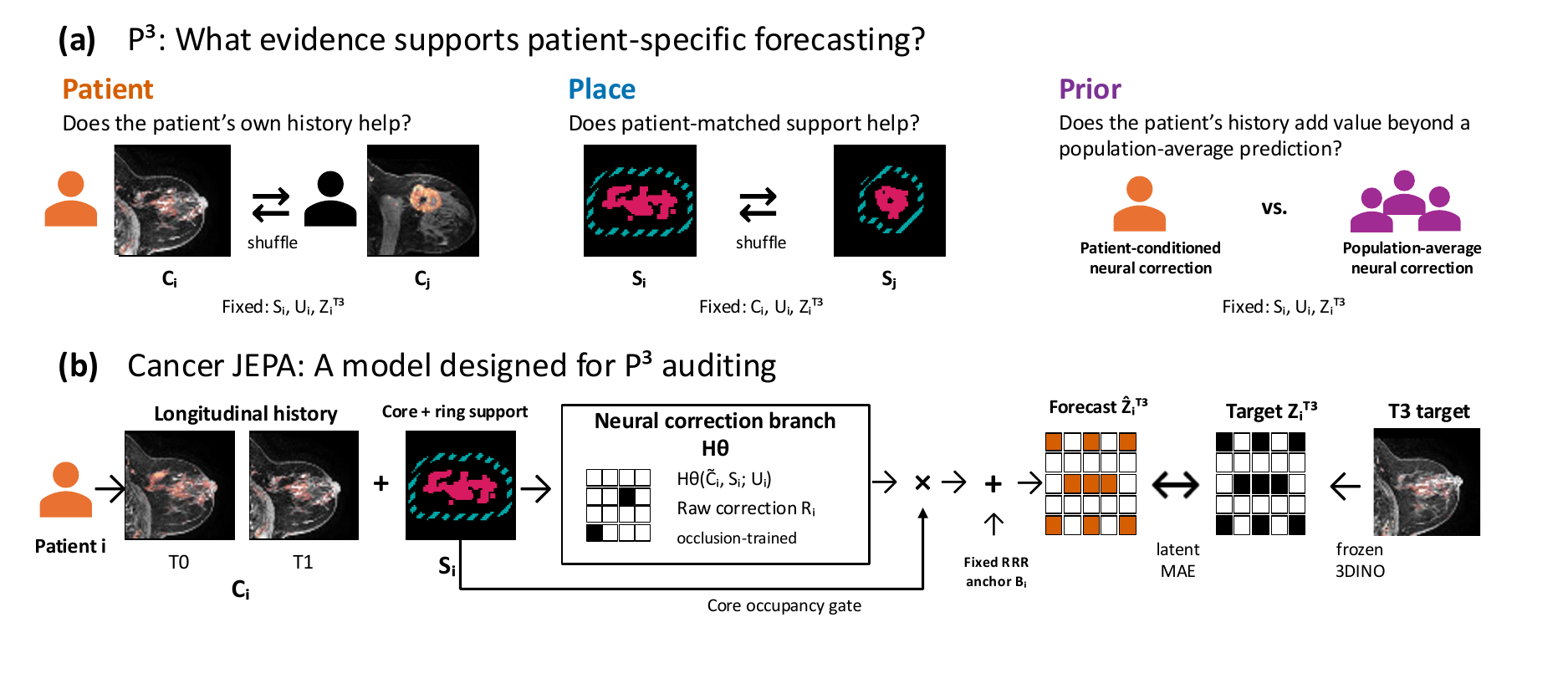}
    \input{figures/figure1_p3_overview_vector_p3first_caption.tex}
    \label{fig:p3_overview}
\end{figure}

Patient, Place, Prior (P$^3$) audits a fixed longitudinal predictor by varying imaging history and externally supplied spatial support independently at inference. \emph{Patient} asks whether the forecast benefits from the patient's own longitudinal history; \emph{Place} asks whether it benefits from patient-matched spatial support; and \emph{Prior} asks whether the patient's history adds predictive value beyond a population-average prediction under matched support and context. P$^3$ applies when history and support can be recombined without changing the prediction target or remaining observed context. Figure~\ref{fig:p3_overview} summarizes the methodology and its Cancer JEPA instantiation.

\subsection{Fixed-patient evaluation}
\label{sec:fixed_patient}

For a fixed patient $i$, write the prediction as
\begin{equation}
    \widehat{Z}_i(C,S)=F(C,S;U_i),
    \label{eq:recipient_fixed}
\end{equation}
and its loss against patient $i$'s future representation as $\ell_i(C,S)$. Here $C$ denotes encoded longitudinal imaging history, $S$ denotes supplied spatial support, and $U_i$ denotes forecast horizon and other observed context. Across all P$^3$ conditions, the model parameters, $U_i$, patient $i$'s future target, and the evaluation procedure remain fixed. Any recipient-specific model component excluded from substitution is also held fixed and reported. Only the source of $C$ or $S$ changes. Let $\pi$ assign each patient an outcome-independent donor such that $\pi(i)\neq i$.

\subsection{Patient and Place: a factorial substitution design}
\label{sec:patient_place}

Patient and Place are evaluated through a $2\times2$ design with matched inputs ($\mathrm{M}$), substituted history ($\mathrm{H}$), substituted spatial support ($\mathrm{L}$, for location), and both inputs substituted ($\mathrm{B}$):
\begin{equation}
\begin{aligned}
    \widehat Z_i^{\mathrm{M}} &= \widehat Z_i(C_i,S_i),
    &\widehat Z_i^{\mathrm{L}} &= \widehat Z_i(C_i,S_{\pi(i)}),\\
    \widehat Z_i^{\mathrm{H}} &= \widehat Z_i(C_{\pi(i)},S_i),
    &\widehat Z_i^{\mathrm{B}} &= \widehat Z_i(C_{\pi(i)},S_{\pi(i)}).
\end{aligned}
\label{eq:factorial_cells}
\end{equation}
Let $\ell_i^{c}$ denote the loss against the unchanged target under condition $c\in\{\mathrm{M},\mathrm{H},\mathrm{L},\mathrm{B}\}$. We define
\begin{equation}
\begin{array}{ll}
    \Delta_{\mathrm{patient}}
        &= \mathbb{E}_i[\ell_i^{\mathrm{H}}-\ell_i^{\mathrm{M}}],\\
    \Delta_{\mathrm{place}}
        &= \mathbb{E}_i[\ell_i^{\mathrm{L}}-\ell_i^{\mathrm{M}}],\\
    \Delta_{\mathrm{joint}}
        &= \mathbb{E}_i[\ell_i^{\mathrm{B}}-\ell_i^{\mathrm{M}}],\\
    \Delta_{\mathrm{interaction}}
        &= \mathbb{E}_i[
            (\ell_i^{\mathrm{L}}-\ell_i^{\mathrm{M}})
            -
            (\ell_i^{\mathrm{B}}-\ell_i^{\mathrm{H}})
        ].
\end{array}
\label{eq:factorial_contrasts}
\end{equation}
$\Delta_{\mathrm{patient}}$ compares the patient's history with donor history under patient-matched support, whereas $\Delta_{\mathrm{place}}$ compares patient-matched with donor support under the patient's history. Positive values indicate lower error with the patient's own input. The joint contrast measures the consequence of replacing both sources, and the interaction asks whether the effect of support depends on which history accompanies it. Because Place varies an externally supplied location, a positive Place contrast establishes beneficial dependence on that support rather than autonomous localization.

\subsection{Population-average prediction}
\label{sec:population_prior}

The Patient contrast compares the patient's history with individual donor histories. Prior instead compares it with a cohort-level reference. Let $\mathcal{Q}_C$ be a distribution of encoded histories defined independently of the evaluated outcomes. We average predictions over this distribution while preserving patient-matched support and observed context:
\begin{equation}
    \widehat{Z}_i^{\mathrm{prior}}
    =
    \mathrm{E}_{\widetilde C\sim\mathcal{Q}_C}
    \left[
        F(\widetilde C,S_i;U_i)
    \right],
    \label{eq:population_prior}
\end{equation}
where prediction-time stochasticity, if present, is included in the expectation. Averaging occurs in prediction space before evaluation. Letting $\ell_i^{\mathrm{prior}}$ denote the loss of this population-average prediction, the Prior contrast is
\begin{equation}
    \Delta_{\mathrm{prior}}
    =
    \mathbb{E}_i[
        \ell_i^{\mathrm{prior}}-\ell_i^{\mathrm{M}}
    ].
    \label{eq:prior_contrast}
\end{equation}
The comparison preserves $S_i$ and $U_i$ and changes only whether the prediction uses $C_i$ or the population reference. Averaging reduces donor-specific variation that can make an individual substitution harmful without showing value beyond a cohort-level pattern. A P$^3$ audit therefore specifies the reference cohort, preserved variables, and prediction-averaging procedure.

\subsection{Evidence interpretation}
\label{sec:p3_inference}

A positive Patient contrast supports beneficial use of the patient's longitudinal imaging history. Evidence of patient-specific predictive value beyond cohort dynamics requires both a positive Patient contrast and a positive Prior contrast under matched support and context. Place is an orthogonal attribution: a positive Place contrast supports beneficial dependence on patient-matched externally supplied spatial support, but it is neither necessary nor sufficient for patient-specific predictive value. The joint and interaction contrasts describe how history and support are combined rather than strengthening the personalization claim.

P$^3$ attributes predictive information use in a fixed forecasting system. Donor substitutions can introduce compatibility shifts between history and support, which is why the population-average reference is evaluated under the patient's unchanged support and context. The resulting statements concern forecast attribution within the observed data distribution; causal treatment effects require a separate identification design.

\section{Cancer JEPA: Lesion-Constrained Latent Forecasting}
\label{sec:cancer_jepa}

Guided by P$^3$, we design Cancer JEPA so that longitudinal imaging history and externally supplied lesion occupancy maps enter the neural correction separately. Cancer JEPA forecasts a future latent MRI representation by adding a lesion-constrained neural correction to a low-complexity reduced-rank regression (RRR) baseline. The construction proceeds from the latent target and RRR baseline to the masked neural correction, its spatial constraint, the occlusion-based objective, and the corresponding P$^3$ comparisons.

\subsection{Longitudinal latent-state forecasting}
\label{sec:cancer_jepa_forecasting}

Let $X_i^{t_0}$, $X_i^{t_1}$, and $X_i^{t_f}$ denote volumetric MRI acquired from patient $i$ at $t_0<t_1<t_f$. A frozen volumetric encoder $E_\phi$ maps every examination into a common spatial representation,
\begin{equation}
    Z_i^t=E_\phi(X_i^t),
    \qquad
    Z_i^t\in\mathbb{R}^{C\times H\times W\times D}.
    \label{eq:cancer_jepa_encoder}
\end{equation}
The frozen encoder defines the common latent space in which earlier and future examinations are compared. Cancer JEPA follows the JEPA principle of predicting a target representation from encoded context rather than reconstructing pixels, but it neither updates the image encoder nor maintains an exponential-moving-average target encoder \citep{lecun2022path,assran2023ijepa,bardes2024vjepa}.

The model receives the latest latent representation, its change from the preceding examination, core and ring lesion occupancy maps, and observed context:
\begin{equation}
\begin{split}
    C_i & = \left(Z_i^{t_1},\,\Delta Z_i\right),
    \qquad \Delta Z_i=Z_i^{t_1}-Z_i^{t_0},\\
    S_i & = \left(O_i^{\mathrm{core}},\,O_i^{\mathrm{ring}}\right),
    \qquad U_i=\left(\Delta t_i,\,a_i\right),
\end{split}
\label{eq:cancer_jepa_inputs}
\end{equation}
where $O_i^{\mathrm{core}}$ and $O_i^{\mathrm{ring}}$ are fractional lesion occupancies derived only from examinations available at prediction time, $\Delta t_i=t_f-t_1$ is the forecast horizon, and $a_i$ contains observed acquisition or clinical variables. The task is to learn $F_\theta(C_i,S_i,U_i)\approx Z_i^{t_f}$. The resulting one-step transition is observational; $a_i$ provides context rather than an identified causal intervention.

\subsection{Low-complexity baseline prediction}
\label{sec:cancer_jepa_anchor}

Strong continuity between serial MRI examinations makes persistence a competitive predictor. We therefore construct a leakage-controlled RRR baseline that combines shrinkage toward a training-partition mean with change patterns learned through a low-rank linear map. Its shrinkage component is
\begin{equation}
    Z_i^{\mathrm{shrink}}
    =
    \alpha Z_i^{t_1}
    +(1-\alpha)\mu_{\mathrm{fit}}^{t_f},
    \label{eq:cancer_jepa_shrinkage}
\end{equation}
where $\mu_{\mathrm{fit}}^{t_f}$ is the mean future representation in the corresponding model-fitting partition and $\alpha$ is selected without using outcomes from the prediction partition. A regularized reduced-rank map $A_\psi$ then predicts the remaining coarse spatial change from pooled imaging history, lesion occupancy maps, and observed context:
\begin{equation}
    B_i
    =
    Z_i^{\mathrm{shrink}}
    +
    \mathcal{U}\!\left[
        A_\psi\!\left(\mathcal{P}(C_i,S_i,U_i)\right)
    \right].
    \label{eq:cancer_jepa_anchor}
\end{equation}
Here $\mathcal{P}$ constructs a coarse representation and $\mathcal{U}$ restores it to the latent grid. Shrinkage, rank, regularization, standardization, and dimensionality reduction are fitted only within the corresponding training partition. Once constructed, $B_i$ is fixed while the neural correction is trained.

The RRR baseline is patient-conditioned because it uses the patient's longitudinal imaging history. It is distinct from the population-average prediction in Section~\ref{sec:population_prior}: the baseline is part of Cancer JEPA, whereas the latter is a post-hoc P$^3$ control used to test the neural correction.

\subsection{Lesion-constrained neural correction}
\label{sec:cancer_jepa_residual}

To prevent the neural branch from learning a position-wise copy of the latest latent representation, Cancer JEPA occludes parts of the encoded imaging history. Let $M_i\in\{0,1\}^{H\times W\times D}$ be an occlusion mask, with $M_{iq}=1$ denoting a hidden token $q$. The occluded inputs are
\begin{equation}
\begin{split}
    \widetilde Z_i^{t_1}
    &=
    (1-M_i)\odot Z_i^{t_1}+M_i\odot m,\\
    \widetilde{\Delta Z}_i
    &=
    (1-M_i)\odot\Delta Z_i,
\end{split}
\label{eq:cancer_jepa_masking}
\end{equation}
where $m$ is a learned mask token and the occlusion mask is broadcast across latent channels. The current representation is replaced by $m$ at hidden locations, while the corresponding longitudinal change is set to zero.

From the masked history, core and ring occupancy, fixed spatial coordinates, and observed context, the neural branch predicts a latent correction. A pointwise projection first maps these inputs to a compact hidden field, condition-modulated three-dimensional blocks exchange local information, and a pointwise head produces the correction. Forecast horizon and observed context modulate the spatial blocks:
\begin{equation}
    R_\theta(C_i,S_i,U_i;M_i)
    =
    s(\Delta t_i)\,
    H_\theta\!\left(
        \widetilde Z_i^{t_1},
        \widetilde{\Delta Z}_i,
        S_i,
        p;
        g(U_i)
    \right),
    \label{eq:cancer_jepa_raw_residual}
\end{equation}
Here $p$ denotes fixed spatial coordinates, $g$ embeds the non-spatial context, and
$s(\Delta t)=\min\{\log(1+\Delta t/30)/\log 7,\,3\}$ is a fixed monotone scale for forecast intervals measured in days. Cancer JEPA adds the neural correction to the RRR baseline only within the supplied lesion core:
\begin{equation}
    \widehat Z_i^{t_f}(M_i)
    =
    B_i+
    O_i^{\mathrm{core}}\odot R_\theta(C_i,S_i,U_i;M_i).
    \label{eq:anchored_prediction}
\end{equation}
This multiplicative gate imposes an exact spatial constraint. For every token $q$,
\begin{equation}
    O_{iq}^{\mathrm{core}}=0
    \Longrightarrow
    \widehat Z_{iq}^{t_f}(M_i)=B_{iq}.
    \label{eq:support_invariant}
\end{equation}
Ring occupancy informs the neural correction as surrounding context, whereas core occupancy determines where the correction can change the RRR baseline. The model therefore depends on externally supplied lesion localization rather than performing autonomous lesion localization.

\subsection{Occlusion-based latent prediction objective}
\label{sec:cancer_jepa_objective}

The neural correction is optimized only where imaging-history tokens are occluded and the supplied core has non-zero occupancy. Let $\mathcal{I}_M=\{i:\sum_q M_{iq}O_{iq}^{\mathrm{core}}>0\}$ denote the eligible patients in a batch. With $\rho$ denoting channel-averaged smooth-$\ell_1$ error, the objective first normalizes within patient and then averages across patients:
\begin{equation}
    \mathcal{L}_{\mathrm{JEPA}}
    =
    \frac{1}{|\mathcal{I}_M|}
    \sum_{i\in\mathcal{I}_M}
    \frac{
        \sum_q M_{iq}O_{iq}^{\mathrm{core}}
        \rho\!\left(
            \widehat Z_{iq}^{t_f}(M_i)-Z_{iq}^{t_f}
        \right)
    }{
        \sum_q M_{iq}O_{iq}^{\mathrm{core}}
    }.
    \label{eq:cancer_jepa_objective}
\end{equation}
Occlusion masks are sampled at multiple coverage levels. At inference, predictions are averaged over the same prespecified occlusion-mask distribution $\mathcal{Q}_M$,
\begin{equation}
    \widehat Z_i^{t_f}
    =
    \mathrm{E}_{M\sim\mathcal{Q}_M}
    \left[
        \widehat Z_i^{t_f}(M)
    \right],
    \label{eq:cancer_jepa_mask_ensemble}
\end{equation}
which reduces dependence on any single occlusion pattern. Training and inference therefore use the same family of history occlusions while the neural correction remains confined to the supplied lesion core.

\subsection{Patient, Place, and Prior in Cancer JEPA}
\label{sec:cancer_jepa_p3}

In Cancer JEPA, Patient replaces the T0/T1 imaging history used by the neural correction, Place replaces the supplied core and ring lesion occupancy maps, and Prior averages neural corrections over training-cohort histories while preserving the patient's lesion occupancy maps and observed context. The future target, RRR baseline, forecast horizon, evaluation procedure, and occlusion-mask distribution remain fixed. The comparisons therefore concern information used by the neural correction rather than every patient-conditioned component of the complete forecast.

The population-average control evaluates encoded training-cohort imaging histories under patient $i$'s lesion occupancy maps and observed context:
\begin{equation}
    \overline R_i^{\mathrm{prior}}
    =
    \mathrm{E}_{\widetilde C\sim\mathcal{Q}_C,\,
                  M\sim\mathcal{Q}_M}
    \left[
        R_\theta(\widetilde C,S_i,U_i;M)
    \right],
    \qquad
    \widehat Z_i^{\mathrm{prior}}
    =
    B_i+
    O_i^{\mathrm{core}}\odot
    \overline R_i^{\mathrm{prior}}.
    \label{eq:cancer_jepa_matched_prior}
\end{equation}
The neural corrections are averaged before applying patient $i$'s core gate and evaluation loss. Comparing the complete model with $B_i$ alone separately measures whether the neural correction improves on the RRR baseline; this model-specific comparison is distinct from Patient, Place, and Prior.

\section{Experiments}
\label{sec:experiments}

\subsection{I-SPY2 cohort and forecasting task}
\label{sec:ispy2_cohort}

I-SPY2 is a multicenter, adaptively randomized phase~II platform trial for patients with high-risk stage~II--III breast cancer. It evaluates investigational regimens in the neoadjuvant setting and uses pathological complete response as its primary clinical endpoint \citep{barker2009ispy2,rugo2016ispy2}. Its public imaging resource contains longitudinal breast DCE-MRI, derived parametric maps, and lesion annotations acquired under a standardized multicenter protocol \citep{li2022ispy2dataset}. MRI is collected at pretreatment (T0), early treatment (T1), between regimens (T2), and before surgery (T3) \citep{onishi2020protocol,li2020ispy2}.

We began with 595 patients having the required longitudinal imaging and spatial annotations. Eight patients were excluded by a prespecified fixed-coverage quality-control rule, yielding 409 training, 88 validation, and 90 test patients. All partitions are patient-disjoint.

Our task forecasts the frozen latent representation of preoperative T3 signal-enhancement-ratio MRI from T0 and T1 imaging history; T2 is not provided as input. The endpoint is neither pathological response, a causal treatment effect, nor a future MRI image, but an observational one-step latent representation. Cancer JEPA is fitted only on the training cohort. Because the validation cohort had already been used during model development, all validation analyses are post-hoc and descriptive rather than independent or confirmatory. The present experiments neither access nor report the test partition.

\subsection{Imaging and model instantiation}
\label{sec:experimental_instantiation}

All images use T0 as the fixed spatial reference. T1 and T3 are rigidly registered to T0, and future examinations never determine crop placement. After a fixed SER intensity transform, a native $64^3$ field centered on the T0 disease bed is resized to $112^3$ and encoded by frozen 3DINO into a $1024\times7\times7\times7$ latent grid \citep{xu2025threedino}. Fractional core and 20-mm ring occupancies are projected from T0 annotations to the same grid. The RRR baseline and Cancer JEPA are fitted only on the 409-patient training partition; all validation and P$^3$ analyses use the resulting frozen model without optimization or model selection.

\subsection{Evaluation and audit protocol}
\label{sec:experimental_protocol}

The primary endpoint is patient-level latent mean absolute error (MAE) within the supplied T0 lesion core. One validation patient has zero core occupancy on the latent grid, giving $N=87$ for the primary analysis; T0 disease bed, 20-mm ring, valid non-disease bed, and full shared grid are secondary regions with $N=88$. The four history--lesion-map conditions use 32 deterministic patient permutations without self-matches and ten fixed occlusion patterns per condition. The patient's future target, RRR baseline, forecast horizon, protocol context, evaluation regions, and occlusion pattern remain fixed. The population-average control evaluates all 409 training-cohort imaging histories under each patient's lesion occupancy maps and observed context, averaging neural corrections before applying the patient's core gate and evaluating the loss.

Errors are first averaged across occlusion patterns and substituted patients within each patient. We then use 10,000 paired patient-level bootstrap replicates to estimate 95\% confidence intervals, treating patients rather than substitutions as the sampling units. All intervals describe post-hoc comparisons on a validation cohort previously used during model development.

\subsection{Incremental value of the neural correction}
\label{sec:anchor_results}

Before attributing forecast differences to patient history or spatial support, we ask whether the neural correction improves its fixed RRR anchor. Table~\ref{tab:anchor_results} shows lower mean MAE for Cancer JEPA in every reported region. The largest estimated improvement occurs within the primary T0 core ($2.15\times10^{-3}$), although its 95\% interval includes zero. The smaller improvements in the disease bed, 20-mm ring, and full grid have intervals above zero. Because the complete prediction equals the RRR anchor wherever the supplied core occupancy is zero, these changes reflect a spatially localized correction. The neural correction therefore yields consistent mean gains across regions, while its incremental value on the primary endpoint remains unresolved.

\begin{table}[t]
    \caption{Validation MAE for Cancer JEPA and its fixed RRR anchor. $\Delta\mathrm{MAE}=\mathrm{MAE}_{\mathrm{RRR}}-\mathrm{MAE}_{\mathrm{JEPA}}$, so positive values favor Cancer JEPA; paired 95\% CIs are reported in $10^{-3}$ units and computed from unrounded errors.}
    \label{tab:anchor_results}
    \centering
    \small
    \begin{tabular}{lrrr}
        \toprule
        Region & Cancer JEPA & RRR & $\Delta\mathrm{MAE}\ (\times 10^{-3})$ [95\% CI] \\
        \midrule
        T0 core & 0.5202 & 0.5224 & $2.15\quad[-0.10,\ 4.25]$ \\
        Disease bed & 0.4714 & 0.4718 & $0.41\quad[0.14,\ 0.70]$ \\
        20-mm ring & 0.4680 & 0.4683 & $0.25\quad[0.08,\ 0.42]$ \\
        Full grid & 0.4266 & 0.4268 & $0.20\quad[0.08,\ 0.34]$ \\
        \bottomrule
    \end{tabular}
\end{table}

\subsection{Predictive attribution with Patient--Place--Prior}
\label{sec:p3_results}

Table~\ref{tab:p3_results} reports the primary-core P$^3$ contrasts. The Patient contrast favors the patient's own longitudinal history over substituted history ($1.48\times10^{-3}$), and the Place contrast favors patient-matched spatial support over substituted support ($0.99\times10^{-3}$); both 95\% intervals lie above zero. The Patient--Place interaction and the Prior contrast have intervals that include zero.

\begin{table}[t]
    \caption{Primary-core P$^3$ contrasts in $10^{-3}$ MAE units. Positive values favor the first condition in each comparison; paired 95\% CIs are descriptive and computed from unrounded errors.}
    \label{tab:p3_results}
    \centering
    \small
    \begin{tabular}{@{}lr@{}}
        \toprule
        Comparison & $\Delta\mathrm{MAE}\ (\times 10^{-3})$ [95\% CI] \\
        \midrule
        Patient: own vs. substituted history & $1.48\quad[0.09,\ 2.88]$ \\
        Place: matched vs. substituted support & $0.99\quad[0.06,\ 1.88]$ \\
        Patient--Place interaction & $0.48\quad[-0.25,\ 1.18]$ \\
        Prior: own history vs. population average & $0.76\quad[-0.67,\ 2.19]$ \\
        \bottomrule
    \end{tabular}
\end{table}

The Patient result shows that the neural correction benefits from the patient's own longitudinal imaging history when the RRR anchor and supplied support are fixed. The Place result shows a separate benefit from patient-matched lesion occupancy maps when imaging history is fixed. Because these maps are externally supplied and core occupancy directly gates the correction, Place establishes dependence on the provided spatial support rather than autonomous lesion localization. The unresolved interaction provides no clear evidence that the value of either input changes with the identity of the other. Crucially, Patient alone is insufficient to establish patient-specific predictive value: the Prior comparison does not show a clear advantage over the population-average neural correction under matched support and context. P$^3$ therefore supports beneficial use of patient history and supplied spatial support in Cancer JEPA, while the present evidence does not establish predictive value beyond cohort-level dynamics. This attribution is distinct from the anchor comparison in Section~\ref{sec:anchor_results}, which asks whether adding the correction improves on the RRR forecast.

\section{Conclusion}
\label{sec:conclusion}

Patient, Place, Prior (P$^3$) reframes personalization in longitudinal medical forecasting as a problem of predictive attribution. It separates the value of a patient's own imaging history, dependence on patient-matched externally supplied spatial support, and improvement over a population-average prediction under matched conditions. We instantiate this methodology in Cancer JEPA, a spatially constrained latent predictor of future breast DCE-MRI representations. In the validation case study, the neural correction is more accurate with the patient's longitudinal history and lesion occupancy maps than with substituted inputs, while its advantage over the matched population-average correction and the complete model's primary-core improvement over the RRR anchor remain unresolved. These results empirically separate patient conditioning, useful patient information, and predictive value beyond cohort dynamics, providing a practical basis for auditable personalization claims in medical world models.

\newpage
\begingroup
\small
\setlength{\bibsep}{4pt plus 1pt}
\bibliographystyle{plainnat}
\bibliography{references}

@article{moor2023gmai,
  title   = {Foundation models for generalist medical artificial intelligence},
  author  = {Moor, Michael and Banerjee, Oishi and Abad, Zahra Shakeri Hossein and Krumholz, Harlan M. and Leskovec, Jure and Topol, Eric J. and Rajpurkar, Pranav},
  journal = {Nature},
  volume  = {616},
  pages   = {259--265},
  year    = {2023},
  doi     = {10.1038/s41586-023-05881-4}
}

@article{zhou2023retfound,
  title   = {A foundation model for generalizable disease detection from retinal images},
  author  = {Zhou, Yukun and Chia, Mark A. and Wagner, Siegfried K. and others},
  journal = {Nature},
  volume  = {622},
  pages   = {156--163},
  year    = {2023},
  doi     = {10.1038/s41586-023-06555-x}
}

@article{wang2026retfoundplus,
  title   = {Time and person sensitive foundation model for disease prediction and risk stratification},
  author  = {Wang, Zheyuan and Zhou, Yukun and Wu, Yilan and others},
  journal = {npj Digital Medicine},
  volume  = {9},
  pages   = {444},
  year    = {2026},
  doi     = {10.1038/s41746-026-02524-6}
}

@article{li2020ispy2,
  title   = {Predicting breast cancer response to neoadjuvant treatment using multi-feature {MRI}: results from the {I-SPY 2} trial},
  author  = {Li, Wen and Newitt, David C. and Gibbs, Jessica and others},
  journal = {npj Breast Cancer},
  volume  = {6},
  pages   = {63},
  year    = {2020},
  doi     = {10.1038/s41523-020-00203-7}
}

@article{barker2009ispy2,
  title   = {{I-SPY 2}: An adaptive breast cancer trial design in the setting of neoadjuvant chemotherapy},
  author  = {Barker, Anna D. and Sigman, Carola C. and Kelloff, Gary J. and Hylton, Nola M. and Berry, Donald A. and Esserman, Laura J.},
  journal = {Clinical Pharmacology \& Therapeutics},
  volume  = {86},
  number  = {1},
  pages   = {97--100},
  year    = {2009},
  doi     = {10.1038/clpt.2009.68}
}

@article{rugo2016ispy2,
  title   = {Adaptive randomization of veliparib--carboplatin treatment in breast cancer},
  author  = {Rugo, Hope S. and Olopade, Olufunmilayo I. and DeMichele, Angela and others},
  journal = {New England Journal of Medicine},
  volume  = {375},
  number  = {1},
  pages   = {23--34},
  year    = {2016},
  doi     = {10.1056/NEJMoa1513749}
}

@article{onishi2020protocol,
  title   = {Impact of {MRI} protocol adherence on prediction of pathological complete response in the {I-SPY 2} neoadjuvant breast cancer trial},
  author  = {Onishi, Natsuko and Li, Wen and Gibbs, Jessica and others},
  journal = {Tomography},
  volume  = {6},
  number  = {2},
  pages   = {77--85},
  year    = {2020},
  doi     = {10.18383/j.tom.2020.00006}
}

@misc{li2022ispy2dataset,
  title        = {{I-SPY 2} Breast Dynamic Contrast Enhanced {MRI} Trial ({ISPY2})},
  author       = {Li, Wen and Newitt, David C. and Gibbs, Jessica and others},
  howpublished = {The Cancer Imaging Archive},
  year         = {2022},
  doi          = {10.7937/TCIA.D8Z0-9T85}
}

@inproceedings{ha2018worldmodels,
  title     = {Recurrent World Models Facilitate Policy Evolution},
  author    = {Ha, David and Schmidhuber, J{"u}rgen},
  booktitle = {Advances in Neural Information Processing Systems},
  volume    = {31},
  pages     = {2451--2463},
  year      = {2018},
  url       = {https://papers.nips.cc/paper/7512-recurrent-world-models-facilitate-policy-evolution}
}

@inproceedings{hafner2019planet,
  title     = {Learning Latent Dynamics for Planning from Pixels},
  author    = {Hafner, Danijar and Lillicrap, Timothy and Fischer, Ian and Villegas, Ruben and Ha, David and Lee, Honglak and Davidson, James},
  booktitle = {Proceedings of the 36th International Conference on Machine Learning},
  series    = {Proceedings of Machine Learning Research},
  volume    = {97},
  pages     = {2555--2565},
  year      = {2019},
  url       = {https://proceedings.mlr.press/v97/hafner19a.html}
}

@article{hafner2025dreamerv3,
  title   = {Mastering diverse control tasks through world models},
  author  = {Hafner, Danijar and Pasukonis, Jurgis and Ba, Jimmy and Lillicrap, Timothy},
  journal = {Nature},
  volume  = {640},
  pages   = {647--653},
  year    = {2025},
  doi     = {10.1038/s41586-025-08744-2}
}

@misc{yang2025mewm,
  title         = {Medical World Model: Generative Simulation of Tumor Evolution for Treatment Planning},
  author        = {Yang, Yijun and Wang, Zhao-Yang and Liu, Qiuping and Sun, Shuwen and Wang, Kang and Chellappa, Rama and Zhou, Zongwei and Yuille, Alan and Zhu, Lei and Zhang, Yu-Dong and Chen, Jieneng},
  year          = {2025},
  eprint        = {2506.02327},
  archiveprefix = {arXiv},
  primaryclass  = {cs.CV},
  doi           = {10.48550/arXiv.2506.02327}
}

@article{laubenbacher2024digitaltwins,
  title   = {Digital twins in medicine},
  author  = {Laubenbacher, Reinhard and Mehrad, Borna and Shmulevich, Ilya and Trayanova, Natalia},
  journal = {Nature Computational Science},
  volume  = {4},
  pages   = {184--191},
  year    = {2024},
  doi     = {10.1038/s43588-024-00607-6}
}

@misc{lecun2022path,
  title  = {A Path Towards Autonomous Machine Intelligence},
  author = {LeCun, Yann},
  year   = {2022},
  note   = {OpenReview position paper, version 0.9.2},
  url    = {https://openreview.net/pdf/315d43ba26f55357a84cec9a7ed15a6610094f79.pdf}
}

@inproceedings{assran2023ijepa,
  title     = {Self-Supervised Learning from Images with a Joint-Embedding Predictive Architecture},
  author    = {Assran, Mahmoud and Duval, Quentin and Misra, Ishan and Bojanowski, Piotr and Vincent, Pascal and Rabbat, Michael and LeCun, Yann and Ballas, Nicolas},
  booktitle = {Proceedings of the IEEE/CVF Conference on Computer Vision and Pattern Recognition},
  pages     = {15619--15629},
  year      = {2023},
  doi       = {10.1109/CVPR52729.2023.01499}
}

@misc{bardes2024vjepa,
  title         = {Revisiting Feature Prediction for Learning Visual Representations from Video},
  author        = {Bardes, Adrien and Garrido, Quentin and Ponce, Jean and Chen, Xinlei and Rabbat, Michael and LeCun, Yann and Assran, Mahmoud and Ballas, Nicolas},
  year          = {2024},
  eprint        = {2404.08471},
  archiveprefix = {arXiv},
  primaryclass  = {cs.CV},
  doi           = {10.48550/arXiv.2404.08471}
}

@article{xu2025threedino,
  title   = {A generalizable {3D} framework and model for self-supervised learning in medical imaging},
  author  = {Xu, Tony and Hosseini, Sepehr and Anderson, Chris and Rinaldi, Anthony and Krishnan, Rahul G. and Martel, Anne L. and Goubran, Maged},
  journal = {npj Digital Medicine},
  volume  = {8},
  pages   = {639},
  year    = {2025},
  doi     = {10.1038/s41746-025-02035-w}
}

@article{damour2022underspecification,
  title   = {Underspecification Presents Challenges for Credibility in Modern Machine Learning},
  author  = {D'Amour, Alexander and Heller, Katherine and Moldovan, Dan and others},
  journal = {Journal of Machine Learning Research},
  volume  = {23},
  number  = {226},
  pages   = {1--61},
  year    = {2022},
  url     = {https://jmlr.org/papers/v23/20-1335.html}
}

@article{geirhos2020shortcut,
  title   = {Shortcut learning in deep neural networks},
  author  = {Geirhos, Robert and Jacobsen, J{"o}rn-Henrik and Michaelis, Claudio and Zemel, Richard and Brendel, Wieland and Bethge, Matthias and Wichmann, Felix A.},
  journal = {Nature Machine Intelligence},
  volume  = {2},
  pages   = {665--673},
  year    = {2020},
  doi     = {10.1038/s42256-020-00257-z}
}

@article{degrave2021covid,
  title   = {{AI} for radiographic {COVID-19} detection selects shortcuts over signal},
  author  = {DeGrave, Alex J. and Janizek, Joseph D. and Lee, Su-In},
  journal = {Nature Machine Intelligence},
  volume  = {3},
  pages   = {610--619},
  year    = {2021},
  doi     = {10.1038/s42256-021-00338-7}
}

@article{brown2023shortcut,
  title   = {Detecting shortcut learning for fair medical {AI} using shortcut testing},
  author  = {Brown, Alexander and Tomasev, Nenad and Freyberg, Jan and Liu, Yuan and Karthikesalingam, Alan and Schrouff, Jessica},
  journal = {Nature Communications},
  volume  = {14},
  pages   = {4314},
  year    = {2023},
  doi     = {10.1038/s41467-023-39902-7}
}

@article{arun2021saliency,
  title   = {Assessing the Trustworthiness of Saliency Maps for Localizing Abnormalities in Medical Imaging},
  author  = {Arun, Nishanth and Gaw, Nathan and Singh, Praveer and others},
  journal = {Radiology: Artificial Intelligence},
  volume  = {3},
  number  = {6},
  pages   = {e200267},
  year    = {2021},
  doi     = {10.1148/ryai.2021200267}
}

@article{litjens2017survey,
  title   = {A Survey on Deep Learning in Medical Image Analysis},
  author  = {Litjens, Geert and Kooi, Thijs and Ehteshami Bejnordi, Babak and Setio, Arnaud Arindra Adiyoso and Ciompi, Francesco and Ghafoorian, Mohsen and van der Laak, Jeroen A. W. M. and van Ginneken, Bram and S\'{a}nchez, Clara I.},
  journal = {Medical Image Analysis},
  volume  = {42},
  pages   = {60--88},
  year    = {2017},
  doi     = {10.1016/j.media.2017.07.005}
}

@article{dinsdale2021unlearning,
  title   = {Deep Learning-Based Unlearning of Dataset Bias for {MRI} Harmonisation and Confound Removal},
  author  = {Dinsdale, Nicola K. and Jenkinson, Mark and Namburete, Ana I. L.},
  journal = {NeuroImage},
  volume  = {228},
  pages   = {117689},
  year    = {2021},
  doi     = {10.1016/j.neuroimage.2020.117689}
}

@article{ouyang2022longitudinal,
  title   = {Self-Supervised Learning of Neighborhood Embedding for Longitudinal {MRI}},
  author  = {Ouyang, Jiahong and Zhao, Qingyu and Adeli, Ehsan and Zaharchuk, Greg and Pfefferbaum, Adolf and Sullivan, Edith V. and Pohl, Kilian M.},
  journal = {Medical Image Analysis},
  volume  = {82},
  pages   = {102571},
  year    = {2022},
  doi     = {10.1016/j.media.2022.102571}
}

@article{gong2024individualised,
  title   = {Individualised Prediction of Longitudinal Change in Multimodal Brain Imaging},
  author  = {Gong, Weikang and Beckmann, Christian F. and Smith, Stephen M.},
  journal = {Imaging Neuroscience},
  volume  = {2},
  pages   = {1--19},
  year    = {2024},
  doi     = {10.1162/imag_a_00215}
}

@article{prosperi2020causal,
  title   = {Causal Inference and Counterfactual Prediction in Machine Learning for Actionable Healthcare},
  author  = {Prosperi, Mattia and Guo, Yi and Sperrin, Matthew and Koopman, James S. and Min, Jae S. and He, Xing and Rich, Shannan and Wang, Mo and Buchan, Iain E. and Bian, Jiang},
  journal = {Nature Machine Intelligence},
  volume  = {2},
  pages   = {369--375},
  year    = {2020},
  doi     = {10.1038/s42256-020-0197-y}
}
\endgroup

\end{document}